\documentclass{article}
\usepackage{spconf,amsmath,graphicx,xcolor, nicefrac, url, hyperref}

\hypersetup{
   unicode=false,          
   pdftoolbar=true,        
   pdfmenubar=true,        
   pdffitwindow=false,     
   pdfstartview={FitH},    
   pdftitle={My title},    
   pdfauthor={Author},     
   pdfsubject={Subject},   
   pdfcreator={Creator},   
   pdfproducer={Producer}, 
   pdfkeywords={keyword1} {key2} {key3}, 
   pdfnewwindow=true,      
   colorlinks=true,       
   linkcolor=blue,          
   citecolor=blue,        
   filecolor=magenta,      
   urlcolor=blue           
}

\title{VoxCeleb Enrichment for Age and Gender Recognition}

\name{Khaled Hechmi$^1$$^,$$^2$, Trung Ngo Trong$^1$, Ville Hautamäki$^1$, Tomi Kinnunen$^1$}
\address{
  $^1$School of Computing, University of Eastern Finland, Finland\\
  $^2$DISCO, Università degli Studi di Milano-Bicocca, Italy}
\begin{document}
%
%
%
%
\maketitle
\begin{abstract}
  VoxCeleb datasets are widely used in speaker recognition studies. 
  Our work serves two purposes. First, we provide speaker age labels and (an alternative) annotation of speaker gender.  Second, we demonstrate the use of this metadata by constructing age and gender recognition models with different features and classifiers. We query different celebrity databases and apply consensus rules to derive age and gender labels. We also compare the original VoxCeleb gender labels with our labels to identify records that might be mislabeled in the original VoxCeleb data. 
  On modeling side, we design a comprehensive study of multiple features and models for recognizing gender and age. Our best system, using i-vector features, achieved an F1-score of 0.9829 for gender recognition task using logistic regression, and the lowest mean absolute error (MAE) in age regression, 9.443 years, is obtained with ridge regression. This indicates challenge in age estimation from in-the-wild style speech data.
  

\end{abstract}
\section{Introduction}
\label{sec:intro}

Since its introduction in 2017, the VoxCeleb corpora  \cite{VoxCeleb1,VoxCeleb2} have become widely used in speech research. The audio data, harvested from YouTube recordings of celebrities, represents vast variation in speakers, speaking styles, and environments. This has facilitated development of new methods to handle such a challenging domain. 
VoxCeleb has most heavily been used in speaker recognition studies~\cite{Lee2019,Ding2020} but other examples include speaker diarization~\cite{Horiguchi2020}, universal speech representation~\cite{Shor2020}, and visual avatar synthesis~\cite{Zakharov20}. The availability of both video and audio data facilitates multimodal person authentication experiments, where speech and face are used jointly~\cite{Chen2020}. 

Nonetheless, speaker identity is only one of many potential paralinguistic attributes. Two other notable ones include gender and age. One interesting property of VoxCeleb is that it contains many within-speaker variations --- including recordings of the same person \emph{in different ages}. Speaker recognition systems can be improved by the use of additional metadata such as gender and age~\cite{Kanervisto2017, Senoussaoui2013}. Further, studies in \cite{GONZALEZHAUTAMAKI20171,GONZALEZHAUTAMAKI2019} indicate negative effect of certain age and gender groups on speaker recognition performance. Therefore, we expect age- and gender-aware speaker recognition systems to be robust in multiple scenarios. Besides the purposes of analyzing or enhancing speaker recognition, age and gender metadata has other use cases --- such as constructing age and gender classifiers that is integrated for downstream application.
Unfortunately, while gender metadata is included in VoxCeleb collections, age information is unavailable. This makes it challenging to address the above tasks on the otherwise versatile VoxCeleb data.



\begin{figure}[!t]
  \centering
  \includegraphics[width=\linewidth]{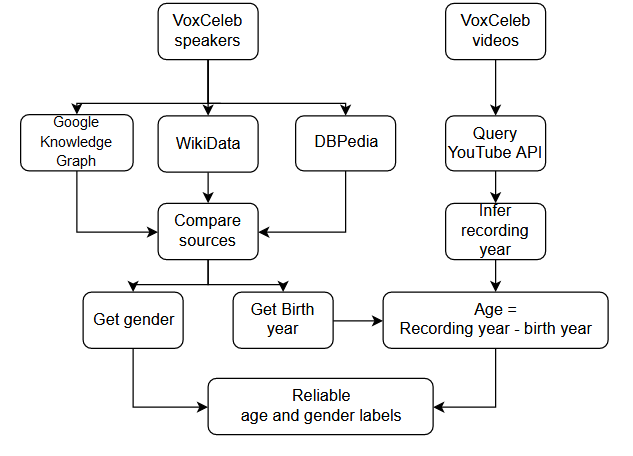}
  \caption{Detail process for extracting age and gender information for each VoxCeleb video using multiple sources.}
  \label{fig:data_enrichment_overview}
\end{figure}

Whether from speech only or from the face image, perceptual experiments have indicated that humans have some capability of estimating the ages of 
persons unknown to them~\cite{Drager2011}. How accurate automatic systems are in this task? How robust automatic age estimation is to different nuisance factors? Age estimation performance is typically evaluated as a MAE between estimated and actual ages. Estimating age from images with recent deep learning methods reaches MAEs in the range 2.17 to 3.34~\cite{Chen2017,Niu2016, Pan2018, Shen2018}. Frontal face and non-occluded face is obviously an easiest to estimate the age from, and the performance can degrade when parts of the face are occluded. With speech some of the challenges include variable length utterances along various technical nuisance factors (e.g. noise and channel) that can mask age-related cues.

Early efforts in estimating age from the speech were based on NIST SRE 2008 and 2010 corpora as the age was included in the metadata. However, these corpora can hardly be considered `in the wild' variety as the participants knew that they are participating in the data collection effort and the topic is controlled by the organizers. 
It is also noteworthy that subjects in the NIST SRE 2008 and 2010 corpora were mostly university students and, therefore, represent limited age range variations. Estimating age from i-vectors yielded MAE of 6.9~\cite{10.5555/2777703.2778083} and neural network back-ends yielded 5.49 MAE for male speakers and 6.35 MAE for females~\cite{FITPUB10971}. Estimating age using senone posterior based i-vectors then yielded 4.7 MAE~\cite{7472637}.

Previous attempts at speaker age estimation use fairly controlled telephony data as the source material. We are interested to find out how age estimation can be performed on the uncontrolled, found data. Thus, the contributions of the our study are two-fold. First, we provide a method for searching age as an additional metadata for the VoxCeleb corpora. It is based on a careful cross-combination and verification of metadata from different celebrity databases as illustrated in Fig. \ref{fig:data_enrichment_overview}. We provide the cleaned metadata for a subset of VoxCeleb videos and for reproduciblity, all metadata and scripts used to produce results found in this paper are to be found here: \url{https://git.io/JYKN4}. Second, we present initial age recognition baselines on this data.

This work was carried out in 2020 when the first author was affiliated with University of Eastern Finland. The authors came later across an independent but closely related work \cite{9414272} that addresses age labeling of VoxCeleb. The key difference between our work and \cite{9414272} is that we assigned age labels based on the videos semantic and people identity, while they trained a facial age estimation model for the labeling task, taking as input the visual frames of the original YouTube videos.



\section{VoxCeleb Metadata}
\subsection{VoxCeleb Corpus}

Since the first official release at Interspeech 2017, the VoxCeleb~1 \cite{VoxCeleb1} dataset has been used in various tasks. The corpus consists of 153516 utterances from 1251 unique speakers, extracted from 22496 different YouTube videos. A year later, VoxCeleb~2 \cite{VoxCeleb2} was released with a larger number of speakers, videos and utterances: the dataset consists of 6112 celebrities, 150480 videos and more than 1 million recordings. Along with speaker identities, the original authors provided gender and nationality information of each person. Concerning metadata, two of our contributions include (i) independent validation of the gender metadata, and (ii) extraction of age metadata. 

\subsection{Metadata Extraction and Cross-Validation}
Gender and age values of each video are obtained by combining and filtering different data sources. First, we used YouTube API to query for the original video ID including its title, description and upload date. Since some of the original videos no longer exist, or are geo-restricted\footnote{Because of our computational environment, the used API credentials are associated with an Italian account and all queries were made from a Finnish IP Address}, we were unable to collect these details for all videos. Nonetheless, $88.5\%$ of the videos' metadata were successfully retrieved.

The next step was to retrieve personal information of all speakers in the corpus. We looked up each speaker's name from multiple independent sources including \emph{Google Knowledge Graph} (GKG), \emph{DBpedia}  and \emph{Wikidata}.  Even if gender and birth date are included in these three databases, there are disagreements among them, which reflects uncertainty in the metadata. For instance, there are celebrities with similar or identical names. As an example, Paolo Ruffini (1765-–1822) was an Italian mathematician and philosopher, while another Paolo Ruffini (1978--) is a present-day Italian actor and presenter. Another source of uncertainty is added by the YouTube video descriptions that may contain wrong information. A \emph{joke} example is the case of Conan O'Brien (a US television host; male) occasionally being mislabeled as Tarja Halonen (former Finnish president; female), due to joke about their likeness.

While we have little means to control the potential errors in the video title or description, we aim for precision and robustness in verifying the gender and birthday metadata. Therefore, we only accept records with unanimous consensus among the three sources, and disregard the rest. Cross-validation among independent sources substantially reduces labeling errors. 
Out of $7365$ unique persons, $5536$ had consistent gender across all three sources ($3437$ male, $2098$ female, and $1$ transgender female). Concerning the birth year, however, only $5157$ speakers had consistent information between GKG, DBPedia and Wikidata.

Since gender labels \emph{are} already provided with the original VoxCeleb 1 and 2 corpora, we were also interested to cross-compare our independently derived labeling with that information. Out of the $5536$ speakers with extracted metadata, the gender labels in the original VoxCeleb metadata agreed in $5494$ cases (99.2\% level of agreement). Of the $42$ disagreed cases, $32$ males were re-labeled as females and $9$ females as males with our approach. The last disagreement is a transgender female (according to our sources), labeled as female in the original VoxCeleb metadata. As a remainder, we treat our gender labeling as the correct ones; our confidence relies on the consensus of the three independent sources.

The last step was to infer the age of each speaker. In an ideal scenario this could be done by computing the difference of the recording date and the speaker's date of birth. In this case, the result will be the exact speaker age. Unfortunately, estimating video recording time is more challenging due to the fact that it could have been recorded at anytime in the past. For simplicity, we focus on the unit of years instead of the exact dates. Even if this discretization introduces potential errors (up to 2 years for some records), it is a reasonable trade-off between complexity and accuracy for our task. To this end, we looked up for references of the upload year --- the only date available --- in the title and the description of the video. If the upload year was found from both fields, the hypothesized recording year of the video was set to that value. Even though title and description are free-text fields, populated by the video uploader, they are part of YouTube ranking factors. Since they provide semantic context about the video for the users, it is likely that they contain related and correct summary information for the given video. Once the recording year is recovered, the speaker age can be trivially inferred as the difference of the recording year and the speaker's birth year. For training, videos with the upload year referenced only in the title were considered as an valid attribute. This was deemed useful for increasing the number of training observations available as $1106$ additional celebrities were added, even though they were never part of the test set considering their lower reliability.

With this approach, we inferred age for $2817$ speakers, and overall $6601$ unique \texttt{<YouTubeID, VoxcelebID, age>} triplets. Note that each person can have multiple different age values specific to each video and age distribution for these $4821$ unique \texttt{<Speaker, Age>} pairs can be found in Table \ref{tab:age_distribution} . For instance, Tom Cruise was interviewed in 2005, 2008 and 2014. Thus, he appeared in (at least) three recordings when he was 24, 27 and 33 years old. For any speaker that occurred in multiple videos but with different age entries, we randomly select only single age value for the speaker. This is a deliberate strategy to reduce over-presentation of certain popular speakers in both training and testing data. In this study, we were not interested in modelling how a given, fixed person's voice changes through time. In other words, we prevent scenarios where the model will always predict the same age for a given speaker as we wish our predictors to focus only on population features.

\begin{table}[h]
  \begin{center}
  \caption{Age distribution for labeled speakers}
  \label{tab:age_distribution}
  \begin{tabular}{|c|c|}
  \hline
  Age interval & Number of speakers \\
  \hline
    (0, 10]   &      1 \\ \hline
    (10, 20]  &    112 \\ \hline
    (20, 30]  &   1204 \\ \hline
    (30, 40]  &   1287 \\ \hline
    (40, 50]  &    926 \\ \hline
    (50, 60]  &    702 \\ \hline
    (60, 70]  &    457 \\ \hline
    (70, 80]  &    122 \\ \hline
    (80, 92]  &    10  \\ \hline
    
  \end{tabular}
  \end{center}
\end{table}

\section{Methodology}

\subsection{Feature processing}

A number of different feature representations were evaluated for the age and gender recognition tasks. As suggested in \cite{probing-xvec}, i-vector \cite{i-vector} and x-vectors \cite{x-vector} contain high amount of relevant information of speakers, including gender and channel variations. Our first choice is to use these recording-level features. Both the i- and x-vector extractors were trained using mel-frequency cepstral coefficient (MFCC) features for \emph{speaker recognition}, rather than age or gender recognition. The reason is that speaker IDs are available for a much larger collection than the age and gender labels explained above. The processes were performed by \texttt{ASVTorch}~\cite{ASVTorch}, a Python 
wrapper built on top of Kaldi and PyTorch. 

Additionally, for the age regression task two different frame-level feature sets were used. The first one consists of logarithm of mel-frequency filter outputs, the latter the same but with an additional discrete cosine transform (DCT) --- i.e., MFCCs. The settings are exactly the same as for i/x-vector models. Specifically, we use 25 ms frames, 30 mel filters and 24 MFCCs. The same processing is done also for augmented recordings, obtained by mixing the MUSAN \cite{MUSAN-dataset} dataset with the original tracks to get 4 additional tracks per utterance. 

Due to differing recording lengths, it was necessary to unify the length of MFCC and log-mel power features. Non-speech frames were excluded, then,
we fix the desired length on the temporal axis to $200$ frames (i.e. 2 seconds) and sample random chunk of data spanning frames $[k:k+199]$ where the start index $k$ is drawn at random in range
$[0,N_\text{f}-200]$ and $N_\text{f}$ being the number of frames in the utterance.
During training, $k$ is changed at each epoch in order to reduce the chance of overfitting. However for the test set the start index is picked before the training phase and is never changed in order to guarantee the reproducibility of experiments.

\subsection{Predictive algorithms}

In the \textbf{gender recognition} task, we evaluate both shallow and deep models to investigate the impact of model size while ensuring feasible training times. Two representative back-end models where used with the recording-level i- and x-vector features: logistic regression and feed forward neural networks, with 2 and 4 hidden dense layers, having all 512 neurons apart from the output node.

For the \textbf{age regression} task, in turn, we consider neural networks-based models, such as feed-forward neural networks with 2 hidden dense layers made of 512 neurons each and convolutional neural networks (1-dimensional and 2-dimensional) with optional residual connections, as well as linear, LASSO and ridge regression.

LASSO fitting solves $\min \{ {({Y} - \hat{Y})}^2 +\lambda \sum_{i=1}^{D}|\hat{\beta}_i| \}$, where $Y$ and $\hat{Y}$ are the ground truth and prediction vectors, respectively, $\hat{\beta}_i$ is the regression coefficient of the \textit{i}-th feature, $D$ is the number of  features (here, $D=512$) and $\lambda$ is a regularization parameter that encourage sparsity constraint on the parameters. Ridge regression, in turn, optimizes $\min \{ {({Y} - \hat{Y})}^2 +\lambda \sum_{i=1}^{D}{\hat{\beta}_i}^2\}$, the additional regularization forces the parameters to be small, hence, less sensitive to the change in input features and improving generalization.

With MFCCs and power mel-spectrum inputs, most of our models have a single input and a single output. However, some of our CNNs contain multiple inputs and multiple outputs as well. Particularly, we consider both continuous (e.g. 25.0) and quantized (e.g. `age category 25-30') data for the age regression task. The model architecture follows the design of siamese neural networks with shared parameters apart from the output layer. The key idea is that the quantization would balancing the label distribution among age groups. 
Most of the models trained for the age regression task used the mean-squared error (MSE) as loss function: the only exceptions are the multi-input multi-output models, where the loss is $0.1 \times \mathrm{MSE} + 10 \times \mathrm{CE}$, where $\mathrm{CE}$ stands for cross-entropy. The weights $0.1$ and $10$ were chosen empirically to balance the contribution of each loss term.

\section{Experimental Setup}

\subsection{Data split and cross-validation set-up}
In order to avoid data leakage from test set to training set, several precautions were taken. First, for both tasks, the data was split between recordings where age or gender is available and the ones where these attributes are absent. This latter data, consisting of the whole VoxCeleb 1 corpus (1251 speakers), was used for training the x- and i-vector extractors. 

The remaining part (with age and gender labels) is reserved for training and testing the gender and age recognizers. The data is subjected to holdout, with 60\% and 40\% of the data used for training and testing, respectively. \textit{Five-fold cross-validation} is used in feature and model selection, for tuning hyperparameters. Importantly, no speakers are shared between training and validation folds, or between training and test data. 

For the gender recognition task, we ensure that male and female labels had equal amounts at every step, to avoid imbalance and potential gender bias. In the age regression task, instead, train and test distributions are kept close as the test set is obtained by randomly sampling speakers. During testing, for each step of cross-validation, the same number of recordings is used for each speaker to ensure that each speaker contributes equally to the metric of interest. The same principle is followed during training unless otherwise noted. 

\subsection{Evaluation metrics}

For the gender recognition task, we evaluate models using \textit{F1-score}, a commonly used metric in different binary classification tasks. F1 score is preferred over accuracy to avoid favoring models biased towards dominant class. It is computed as $\mathrm{F1} = \frac{2 \times \mathrm{precision} \times \mathrm{recall}}{\mathrm{precision} + \mathrm{recall}}
$, where $\mathrm{precision} = \nicefrac{\mathrm{TP}}{\mathrm{TP} + \mathrm{FP}}$ and $\mathrm{recall} = \nicefrac{\mathrm{TP}}{\mathrm{TP} + \mathrm{FN}}$. Here, $\mathrm{TP}$, $\mathrm{FP}$, and $\mathrm{FN}$ denote the number of true positives, false positives and false negatives, respectively.
The convention adopted for our work represents males as the positive class and females as the negative class. This is an arbitrary convention and does not impact the reported F1 scores. 

Age regression models, in turn, are evaluated using MAE, defined as $\text{MAE}=\frac{1}{N_\text{test}}\sum_{i=1}^{n}|y^{(i)}_\text{true} - y^{(i)}_\text{predicted}|$, where $N_\text{test}$ is the number of test cases while $y^{(i)}_\text{true}$ and $y^{(i)}_\text{predicted}$ denote the ground-truth and predicted age, respectively. No rounding was applied to model predictions before computing this score. During training, mean F1 and mean MAE across folds across test folds were computed in order to identify the best model and training configuration. 

\subsection{Reference values for classification}\label{subsec:classification-reference}

Even if not common in age recognition studies, it is useful to reference classifier predictions to less informative systems, namely, \emph{guessing}. To this end, we consider three different reference values for the MAE metric with different prior knowledge of the test data. The first approach replaces classifier predictions with random predictions drawn (with replacement) from the empirical distribution of test set ages. The process is repeated 100k times and yields an average MAE of $16.330$ years. The second approach is similar but the predictions are drawn from a less informative uniform distribution instead, with the range set according to the minimum and maximum age. This yields an average MAE of $23.212$ years. Our final approach predicts a fixed value \emph{deterministically}. In particular, using  grid search, we found the optimal (lowest MAE) fixed age of 39 years, with the respective MAE of $12.243$ years. 
 
\section{Results}

\subsection{Gender recognition}

All the gender recognition experiments were balanced in terms of class and without data augmentation used in training. As Table~\ref{tab:1} indicates, both x- and i-vectors are highly informative of the gender. This is consistent with the findings in \cite{probing-xvec} and the fact that a linear model (i.e. logistic regression) outperforms deeper networks indicates a linear relation between the embedding vectors and age information. Overall, the difference between the two types of embedding is minor, and the performance is less likely to be affected by channel variation since the accuracy is above 95\% both in our case and in \cite{probing-xvec}. 

\begin{table}[h]
    \begin{center}
    \caption{Gender recognition}
    \label{tab:1}
     \begin{tabular}{|c|c|c|c|c|c|} 
     \hline
     Model & i-vector F1-score & x-vector F1-score \\ [0.5ex] 
     \hline
      FC2 & 0.9823 & 0.9776 \\
      \hline
      FC4 & 0.9805 & 0.9776 \\ 
     \hline
      Logistic reg. & \textbf{0.9829} & 0.9769 \\ 
     \hline
    \end{tabular}
    \end{center}
\end{table}
\subsection{Age regression}

Several different experiments were carried out in the more challenging, age regression task. We begin by comparing linear regression, ridge and LASSO back-ends with both types of embedding features. The training and testing datasets are balanced in the number of utterances selected from each speaker.
The results are reported in Table ~\ref{tab:age_ivec_xvec}. 

\begin{table}[h]
  \begin{center}
  \caption{Age regression, comparison between x-vector vs i-vector based systems with equalized training utterances}
  \label{tab:age_ivec_xvec}
  \begin{tabular}{|c|c|c|}
  \hline
  Model & i-vector MAE & x-vector MAE \\
  \hline
  Linear reg & 10.332 & 11.095 \\
  \hline
  Ridge & 10.402 & \textbf{9.777} \\
  \hline
  LASSO & 10.126 & 9.843 \\
  \hline
  \end{tabular}
  \end{center}
\end{table}
Concerning the back-end, ridge regression yield the best MAE for x-vector, while LASSO is the most promising model for i-vector. We also tried non-linear estimator (deep neural network) which however resulted in overfitting and degraded performance on the validation set. Concerning the features, x-vector yield lower MAEs in most cases.

\begin{table}[h]
  \begin{center}
  \caption{Age regression without balancing the age distribution, comparison between x-vector and i-vector based system}
  \label{tab:age_equal_train_utt}
  \begin{tabular}{|c|c|c|}
  \hline
  Model & i-Vector MAE & x-vector MAE \\
  \hline
  Lin Regr & \textbf{9.443} & 9.962 \\
  \hline
  Ridge & 9.539 & 9.768 \\
  \hline
  LASSO & 9.516  & 9.789 \\
  \hline
  \end{tabular}
  \end{center}
\end{table}

We carried another similar experiment \emph{without} data balancing, to find out whether using all the available training data may help in learning a better representation. Table~\ref{tab:age_equal_train_utt} indicates the robustness of i-vector under the impact of imbalanced age distribution. This is anticipated since x-vector was trained using supervised information (i.e. speaker labels) which is directly connected to the age labels. As a result, the training procedure could incorporate unexpected inductive biases toward certain dominant age groups. 
The best overall result is now obtained with i-vector embedding and linear regression model.

\begin{table}[h]
  \begin{center}
  \caption{Age regression, comparison between end-to-end acoustic systems
  (*: Train set with data augmentation and observations where the upload year is referenced only in the title)}
  \label{tab:age_ivec_acous}
  \begin{tabular}{|c|c|c|}
  \hline
  Model & MFCC MAE & Log Mel MAE \\
  \hline
  CNN 1-D* & \textbf{9.443} & 16.770 \\
  \hline
  CNN 1-D Multi Output* & 9.510  & 13.326 \\
  \hline
  \end{tabular}
  \end{center}
\end{table}

Since neither i- nor x-vectors are specifically designed for age recognition, we made an attempt to learn more informative model by combining frame-level features (here, MFCCs and log-mel spectrograms) with a CNN-based model. Deep learning has been demonstrated as a powerful end-to-end framework that automatically extracting relevant information for various speech processing task \cite{DeepLanguage,DeepSpeech,DeepSpeaker}. Our best convolutional model has three convolutional layers of 30, 60 and 120 filters respectively, and two dense layers of 256 and 128 nodes, before the output node.
Table.~\ref{tab:age_ivec_acous} shows that CNN with MFCC obtain the same MAE obtained by the best i-vector system reported.
We suspect the reason why frame-level features do not lead to better results is related to the amount of training data: while the age regressor (back-end) training set is the same for all the systems in Table.~\ref{tab:age_ivec_acous}, the i-vector extractor is trained using a large offline training data. This may help in reducing bias due to imbalanced age distribution, while data augmentation was used for the end-to-end approaches, these utterances still come from the same set of speakers with biases from dominant age groups. This issue is intensified by strong non-linearity of deep models could easily steer the model toward sub-optimal region.

\begin{figure}[!t]
  \centering
  \includegraphics[width=\linewidth]{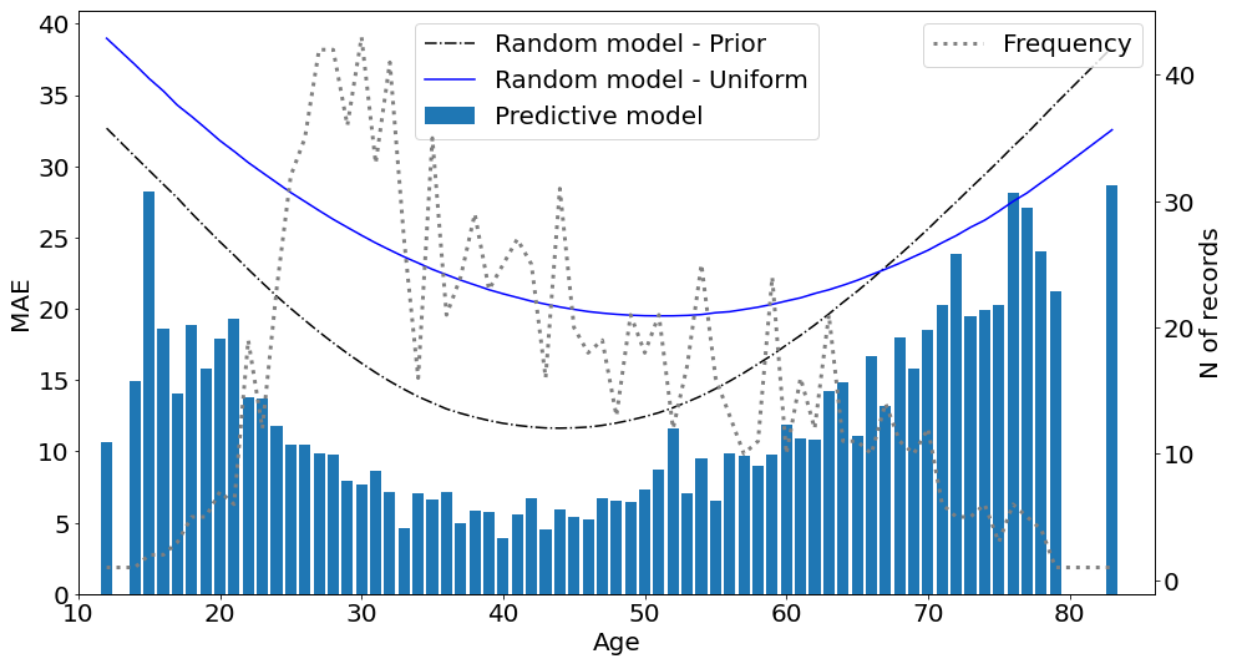}
  \caption{MAE variation across ages for best model.}
  \label{fig:mae_across_age}
\end{figure}

Finally, we contrast our best model (i-vector features with linear regression) with the three references explained in section \ref{subsec:classification-reference}. The model predictions are substantially better than the MAE obtained by these approaches, an indication that the model learns to predict ages. But as Figure \ref{fig:mae_across_age} reveals, the results vary considerably across ages: the most accurate predictions are obtained for the age range $[30\dots55]$ with an MAE as low as $5.5\dots6$ in the range of 40 years.

We suspect the non-uniformity of the age distribution itself to be the primary reason for the non-uniform performance. The number of utterances for 40 year old speakers outnumbers the data available for the other ages and model predictions are confined to a narrower range of $[19\dots68]$ years. This is somewhat surprising considering that humans are usually good at identifying `broad age ranges' --- and certainly young and old people are not easily exchanged for middle age-ones. 
Note also that prediction errors around the median age values are also lower for the reference classifier: by drawing a random number between the minimum and maximum ages yields `better' predictions for the medium ages than near the maximum and minimum ages.

\section{Conclusions}

We have demonstrated how to   
extend the usage of Voxceleb dataset 
for additional tasks like age and gender recognition.
Our pilot experiments indicate high recognition performance in  
the gender recognition task: the F1 score of our best model was 0.9829. 
Age regression, however, turned out a much more complex task. 
The best achieved overall MAE was $9.443$. While substantially lower than baseline values obtained by random guessing or fixed age predictions, the error remains high. Considering high variability of MAE across ages, future study could focus on better data augmentation using self-supervised learning or Bayesian approach that addresses the biases via age distribution prior. We also release the metadata to be used for improving speech processing tasks using the VoxCelebs dataset.

\section{Acknowledgements}

This work has been partially sponsored by Academy of Finland (proj. no. 309629). 

\bibliographystyle{IEEEbib}
\bibliography{strings}

\end{document}